\documentclass{ecai} 
\usepackage{latexsym}
\usepackage{amssymb}
\usepackage{amsmath}
\usepackage{amsthm}
\usepackage{booktabs}
\usepackage{enumitem}
\usepackage{graphicx}
\usepackage{subcaption}
\usepackage{color}
\usepackage{hyperref}
\usepackage[table,xcdraw]{xcolor}
\usepackage[justification=centering]{caption}

\newcommand{\BibTeX}{B\kern-.05em{\sc i\kern-.025em b}\kern-.08em\TeX}

\begin{document}

%%%%%%%%%%%%%%%%%%%%%%%%%%%%%%%%%%%%%%%%%%%%%%%%%%%%%%%%%%%%%%%%%%%%%%%%

\begin{frontmatter}

%%% Use this command to specify your submission number.
%%% In doubleblind mode, it will be printed on the first page.

% \paperid{1319} 

%%% Use this command to specify the title of your paper.

\title{Interpretable Robotic Manipulation from Language}

%%% Use this combinations of commands to specify all authors of your 
%%% paper. Use \fnms{} and \snm{} to indicate everyone's first names 
%%% and surname. This will help the publisher with indexing the 
%%% proceedings. Please use a reasonable approximation in case your 
%%% name does not neatly split into "first names" and "surname".
%%% Specifying your ORCID digital identifier is optional. 
%%% Use the \thanks{} command to indicate one or more corresponding 
%%% authors and their email address(es). If so desired, you can specify
%%% author contributions using the \footnote{} command.

% \author[A]{\fnms{Boyuan}~\snm{Zheng}\orcid{....-....-....-....}}
\author[A]{\fnms{Boyuan}~\snm{Zheng}}
\author[A]{\fnms{Jianlong}~\snm{Zhou}}
\author[A]{\fnms{Fang}~\snm{Chen}} 

\address[A]{University of Technology Sydney}

%%% Use this environment to include an abstract of your paper.

\begin{abstract}
Humans naturally employ linguistic instructions to convey knowledge, a process that proves significantly more complex for machines, especially within the context of multitask robotic manipulation environments. Natural language, moreover, serves as the primary medium through which humans acquire new knowledge, presenting a potentially intuitive bridge for translating concepts understandable by humans into formats that can be learned by machines. In pursuit of facilitating this integration, we introduce an explainable behavior cloning agent, named Ex-PERACT, specifically designed for manipulation tasks. This agent is distinguished by its hierarchical structure, which incorporates natural language to enhance the learning process. At the top level, the model is tasked with learning a discrete skill code, while at the bottom level, the policy network translates the problem into a voxelized grid and maps the discretized actions to voxel grids. We evaluate our method across eight challenging manipulation tasks utilizing the RLBench benchmark, demonstrating that Ex-PERACT not only achieves competitive policy performance but also effectively bridges the gap between human instructions and machine execution in complex environments.
Code, data, and supplementary materials are publicly available at \url{https://anonymous.4open.science/r/ExPERACT}.

%demonstrating its effectiveness in bridging the gap between human instruction and machine execution in complex environments.

\end{abstract}

\end{frontmatter}

%%%%%%%%%%%%%%%%%%%%%%%%%%%%%%%%%%%%%%%%%%%%%%%%%%%%%%%%%%%%%%%%%%%%%%%%

\section{Introduction}

%llm develop, intuitive and problem-solving by using natural language
The exploration of sequential decision-making within the realm of artificial intelligence (AI) is a pivotal research focus, bearing substantial implications for applications such as autonomous driving, gaming strategy, and humanoid robotics \cite{pan2017agile,oh2018self,yu2018one}. Central to this pursuit is imitation learning (IL), which entails learning from demonstrations and emerges as a promising primary approach to effectively address the challenges inherent in sequential decision-making. To enhance the capabilities of intelligent agents, research communities have integrated various forms of auxiliary information, including human preferences, task-specific parameters, and the expertise level of experts \cite{brown2020safe,li2017infogail,beliaev2022imitation}. Notably, advancements in large language models have established natural language as a preferred method among researchers, owing to its intuitiveness and provision of a more universally applicable means for instructing agents. Diverse methodologies proposed in studies such as \cite{stepputtis2020language,roh2020conditional,zhou2021inverse} are aimed at training single-task agents, utilizing the conventional state-action pair as input and supplementing it with natural language as auxiliary information.

%language-conditioned multitask model: importance, reuse, connection between sub-task
% As machine learning techniques continue to advance, the performance of intelligent agents in single tasks has seen gradual improvement, prompting researchers to anticipate proficient multitask domain performance. The designed intelligent agents acquire knowledge from offline demonstrations of various tasks and appropriately interact with the environment during test time. The incorporation of natural language is anticipated to fortify the connection between tasks, leading to an improved ability to reuse acquired skills for similar tasks. Research like PERACT \cite{shridhar2023perceiver} and BC-Z \cite{jang2022bc} demonstrate that integrating natural language into multitask models significantly impacts the success rate of tasks.
As machine learning techniques progress, there has been a gradual enhancement in the performance of intelligent agents across single tasks, instilling anticipation among researchers for adept multitask domain performance. These intelligent agents are designed to acquire knowledge from offline demonstrations spanning various tasks and appropriately interact with the environment during testing time. The integration of natural language holds promise in strengthening the interconnections between tasks, thereby fostering an enhanced ability to reuse acquired skills across similar tasks. Studies such as PERACT \cite{shridhar2023perceiver} and BC-Z \cite{jang2022bc} exemplify how the incorporation of natural language into multitask models significantly influences task success rates, underscoring its potential impact in advancing the field.

%using language is non-trivial: 
%           implicit subtasks caused by common sence
%           unsupervised & language is abstract and not including the subtask division image
%           devisity of the instruction
% However, training a language-conditioned multitask model is non-trivial.
% Firstly, language instructions often fall short in fully elucidating tasks due to inherent limitations in common sense reasoning. For example, an instruction like ``Open drawer" tends to overlook steps such as targeting and gripping the drawer handle, impeding the summarization and reusability of implicit sub-tasks for comparable activities. Moreover, the inherently abstract nature of language provides limited information about subtask divisions and lacks ground truth for potential skills, thereby making the learning process unsupervised. Furthermore, task instructions may exhibit variability among individuals, and the diversity in expression introduces intricacies into the training of a multitask model. Additionally, natural language offers a means for humans to probe the model and enhance interpretability—an aspect that has received limited attention in the realm of multitask model research.
However, training a language-conditioned multitask model poses significant challenges. Firstly, language instructions often fail to fully elucidate tasks due to inherent limitations in common-sense reasoning. For instance, instructions like ``Open drawer" may overlook crucial steps such as targeting and gripping the drawer handle, thereby hindering the summarization and reusability of implicit sub-tasks for comparable activities. Moreover, the inherently abstract nature of language provides limited information about subtask divisions and lacks ground truth for potential skills, rendering the learning process unsupervised. Furthermore, task instructions may vary among individuals, introducing complexities into the training of multitask models due to the diversity in expression. Additionally, while natural language provides a means for humans to probe the model and enhance interpretability, this aspect has garnered limited attention in multitask model research.

%intro my method
% To tackle the aforementioned challenges and further explore explanability, we introduce a novel method named Ex-PerAct. This approach employs a hierarchical structure and incorporates multitask imitation learning to glean insights from language-conditioned offline demonstrations. Ex-PerAct comprises a skill predictor and a policy network. The skill predictor condenses skills from diverse tasks into a discrete skill codebook, providing a comprehensive summary of subdivided demonstration snippets. This model serves as a crucial link connecting human-understandable natural language to digital skill vectors, facilitating the reuse of these condensed skills in analogous tasks. Vector Quantization (VQ) is employed to cluster these skill vectors in an unsupervised manner \cite{van2017neural}. As for the policy network, we leverage the state-of-the-art PerAct \cite{shridhar2023perceiver}, a multitask Behavior Cloning agent renowned for its competitive performance across a broad spectrum of manipulation tasks by representing observations and actions as 3D voxels instead of 2D image pixel.
To address the aforementioned challenges and delve deeper into explainability, we propose a novel method named Ex-PerAct. This approach employs a hierarchical structure and integrates multitask imitation learning to acquire reusable skills from language-conditioned offline demonstrations. Ex-PerAct comprises two transformer-based models in sequence. The top-level model condenses skills from diverse tasks into discrete skill codes, providing a comprehensive summary of segmented demonstration snippets. This model acts as a vital bridge connecting human-understandable natural language to digital skill vectors, thereby facilitating the reuse of condensed skills in analogous tasks. To cluster these skill vectors in an unsupervised manner, we employ Vector Quantization (VQ) \cite{van2017neural}. As for the bottom-level model, we leverage the state-of-the-art PerAct \cite{shridhar2023perceiver}, a multitask Behavior Cloning agent renowned for its competitive performance across a wide array of manipulation tasks. PerAct represents observations and actions as 3D voxels rather than 2D image pixels, contributing to its superior capabilities. We conduct experiments on eight manipulation tasks in the challenging benchmark RLBench \cite{james2020rlbench} with multiple baselines and ablations. The results demonstrate that Ex-PERACT achieves better policy performance in almost all tasks while providing an interpretable way for humans to probe the relationship between language instructions and the agent's decision-making process.

%contribution
We highlight the primary contributions of this paper as follows:
% We introduced a hierarchical imitation learning method called Ex-PerAct which leverages various modalities such as 3D voxels and language instruction, and achieves competitive tasks.
% We exhibit Ex-PerAct could extract reusable skills which is beneficial in multi-task robotic manipulation domain.
% Ex-PerAct also bridges human understandable natural language with machine usable vectors and facilitates interpretability across behaviour, demonstations and natural language.
\begin{itemize}
    \item We present Ex-PerAct, a hierarchical imitation learning method that effectively integrates diverse modalities, including 3D voxels and language instructions, thereby achieving competitive performance across tasks.
    \item Our approach showcases the capacity of Ex-PerAct to extract reusable skills across tasks, offering significant advantages in the realm of multitask robotic manipulation. 
    \item Ex-PerAct forges a crucial link between human-understandable natural language and machine-usable vectors, augmenting interpretability across behavioral patterns, and language instructions.
\end{itemize}

\section{Related Work}
% \subsection{language-conditioned imitation learning}
\subsection{Language-conditioned imitation learning in robotic manipulation}
% %mention a little bit more manipulation
% Utilizing natural language for training intelligent agents has been an evolving trend, not limited to recent years. Chen and Mooney presented a framework that facilitates the learning of a semantic parser, aligning instructions with the world state in navigation tasks \cite{chen2011learning}. The advent of transformer-based models has led to the ubiquity of language-conditioned datasets in recent research. Pioneering studies, such as \cite{jang2022bc,karamcheti2023language,lynch2020language,shridhar2023perceiver,guhur2023instruction}, seamlessly integrate this modality with imitation learning, addressing more intricate continuous manipulation tasks compared to early investigations. 
The integration of natural language into the training of intelligent agents has emerged as a prominent trend, with roots extending beyond recent years. Chen and Mooney introduced a framework facilitating the learning of a semantic parser, aligning instructions with the world state in navigation tasks \cite{chen2011learning}. The rise of transformer-based models has ushered in a proliferation of language-conditioned datasets in contemporary research. Pioneering investigations such as \cite{jang2022bc,karamcheti2023language,lynch2020language,shridhar2023perceiver,guhur2023instruction} seamlessly incorporate this modality with imitation learning, tackling more intricate continuous manipulation tasks compared to earlier studies.

% Significantly, Nair et al. annotated an offline dataset with a crowd-sourced natural language label. They developed method called LOReL which recovers a language-conditioned reward function represented by a simple classifier, specifically designed for integration into multi-task reinforcement learning \cite{nair2022learning}. This annotated dataset has subsequently become instrumental in numerous subsequent investigations focusing on language-conditioned robotic manipulation, such as \cite{garg2022lisa}. Simultaneously, Shridhar et al. introduced CLIPORT, a method leveraging natural language to attain commendable performance levels in both simulated and real-world domains \cite{shridhar2022cliport}. 
Noteworthy among recent advancements is the work by Nair et al., who annotated an offline dataset with crowd-sourced natural language labels. They introduced LOReL, a method that recovers a language-conditioned reward function represented by a simple classifier, specifically tailored for integration into multi-task reinforcement learning \cite{nair2022learning}. This annotated dataset has since played a pivotal role in numerous subsequent investigations focusing on language-conditioned robotic manipulation, exemplified by \cite{garg2022lisa}. Concurrently, Shridhar et al. introduced CLIPORT, a method leveraging natural language to achieve commendable performance levels in both simulated and real-world domains \cite{shridhar2022cliport}.

% Despite the extensive literature in this field, there is a limited amount of research investigating the potential enhancement of explainability when natural language is incorporated. LISA, as proposed by Grag et al., leverages a hierarchical IL framework that learns a interpretable skill abstraction from a language conditioned dataset using the cutting-edge transformer-based model, elucidating the correlation between natural language and acquired skills. However, LISA introduced additional parameters such as horizon, and the model's performance strongly rely on these parameters. The evaluation in multitasks related to manipulation is also notably restricted, and its performance in established benchmarks like RLBench is also limited \cite{garg2022lisa}.
Despite the breadth of literature in this domain, research into the potential enhancement of explainability through the incorporation of natural language remains relatively limited. LISA, as proposed by Garg et al., employs a hierarchical imitation learning framework that learns interpretable skill abstractions from a language-conditioned dataset using state-of-the-art transformer-based models, elucidating the correlation between natural language and acquired skills \cite{garg2022lisa}. However, LISA introduces additional parameters such as horizon, and the model's performance is heavily reliant on these parameters. Moreover, its evaluation in multitasks related to manipulation is notably constrained, and its performance in established benchmarks like RLBench is also limited.

\subsection{Hierarchical imitation learning}

% The hierarchical structure has proven beneficial in numerous studies on imitation learning. A notable advantage is its ability to leverage benefits from different models. For instance, Le et al. introduced a hierarchical framework for both imitation learning (IL) and reinforcement learning (RL). Their study demonstrated that employing various combinations of IL and RL at different levels could significantly reduce expert effort and exploration costs. Compared to standard IL, their proposed framework proved to be more label-efficient \cite{le2018hierarchical}.
The hierarchical structure has emerged as a beneficial paradigm in numerous studies on imitation learning, offering various advantages derived from its ability to harness benefits from different models. One notable advantage is exemplified by Le et al., who introduced a hierarchical framework encompassing both imitation learning (IL) and reinforcement learning (RL). Their study showcased that employing different combinations of IL and RL at distinct levels could notably mitigate expert effort and exploration costs, rendering their proposed framework more label-efficient compared to standard IL \cite{le2018hierarchical}.

% Another strength of hierarchical frameworks lies in their ability to facilitate the learning of separate sub-task policies for complex tasks, resulting in performance improvements. Hierarchical option frameworks represent a prevalent approach for learning sub-task policies, categorized into those trained with ground truth option labels \cite{zhang2021explainable,leech2019explainable} and those using alternative forms of supervision \cite{sharma2018directed,wang2022hierarchical,garg2022lisa}. For the former type, an example is FIST, proposed by Hakhamaneshi et al., learns a inverse dynamic model to extract generalizable skills offline and achieves competitive performance in long-horizon tasks under few-shots setting\cite{hakhamaneshi2022hierarchical}.
% In the latter category, Zhang and Paschalidis investigated the Expectation-Maximization approach to hierarchical IL from a theoretical perspective and developed the first convergence guarantee algorithm that only observes primitive state-action pairs \cite{zhang2021provable}.
Furthermore, hierarchical frameworks excel in facilitating the learning of separate sub-task policies for complex tasks, leading to performance enhancements. Hierarchical option frameworks represent a prevalent approach for learning sub-task policies, categorized into those trained with ground truth option labels \cite{zhang2021explainable,leech2019explainable} and those utilizing alternative forms of supervision \cite{sharma2018directed,wang2022hierarchical,garg2022lisa}. For instance, FIST, proposed by Hakhamaneshi et al., falls under the former category, employing an inverse dynamic model to extract generalizable skills offline and achieving competitive performance in long-horizon tasks under few-shot settings \cite{hakhamaneshi2022hierarchical}. Conversely, in the latter category, Zhang and Paschalidis explored the Expectation-Maximization approach to hierarchical IL from a theoretical standpoint, devising the first convergence guarantee algorithm that solely observes primitive state-action pairs \cite{zhang2021provable}.

% In addition to the improvement in performance, hierarchical IL also enhances explainability which will be introduced in the following subsection. In general, the high-level options can serve as a bridge between human understandable knowledge and the model's performance, acting as a bottleneck to facilitate a clearer understanding. Our work also follows the similar idea but highlight the explainability on language over the input demonstrations.
In addition to performance improvements, hierarchical IL also enhances explainability, as will be further elucidated in the subsequent subsection. Broadly, the high-level options serve as a bridge between human-understandable knowledge and the model's performance, acting as a bottleneck to facilitate clearer comprehension. Our work similarly follows this principle but emphasizes explainability regarding language over the input demonstrations.

\subsection{Explainable imitation learning}

% Investigating the exlainability of IL methods becomes more prevalent in recent years. Early research such as \cite{li2017infogail}
% tries to learn interpretable and meaningful representation to infer latent structure of input demonstrations, but it fails to deeper investigate the improvement in explainability for the non-technical. Later in 2020, Pan et al. proposed a method called xGAIL and claimed that xGAIL is the first explainable GAIL framework. With prevalence of explainable AI, more and more research that combines XAI and IL emerges. 
The exploration of the explainability of IL methods has gained traction in recent years, with a notable shift towards enhancing interpretability for non-technical audiences. Early research, proposed in \cite{li2017infogail}, aimed to learn interpretable and meaningful representations to infer the latent structure of input demonstrations, albeit without delving deeply into improving explainability for non-technical users. Subsequently, in 2020, Pan et al. introduced xGAIL, claiming it as the first explainable GAIL framework. As explainable AI gains prevalence, an increasing number of studies are emerging that combine eXplainable AI (XAI) with IL methodologies.

% Leech proposed an intrinsically interpretable IL method, which leverages hierarchical learning framework and combines IL with logical automata, representing problems as compact finite state automata with human-interpretable logic states \cite{leech2019explainable}.
% Bewley et al. (2020) also proposed an explainable IL method using the interpretable model, they modeled the behavior policy of a trained black-box agent using a decision tree generated from analyzing its input-output statistics \cite{bewley2020modelling}.
Leech proposed an intrinsically interpretable IL method that leverages a hierarchical learning framework and integrates IL with logical automata. This approach represents problems as compact finite state automata with human-interpretable logic states \cite{leech2019explainable}. Similarly, Bewley et al. (2020) introduced an explainable IL method employing interpretable models, wherein they model the behavior policy of a trained black-box agent using a decision tree generated from analyzing its input-output statistics \cite{bewley2020modelling}.

% Besides the intrinsically interpretable approach, post-hoc explanation is also prevalent in research community \cite{wang2022hierarchical,xie2022towards,garg2022lisa,jiang2023interpretable}. Zhang et al. (2021) leveraged a hierarchical framework to achieve post-hoc explanation, decomposing the complex task and explaining the model's decision-making process and the causes of failure \cite{zhang2021explainable}. Wang et al. proposed subgoal conditioned hierarchical IL to mimic doctors' behavior and achieve to provide more explainable post-hoc recommendation in dynamic treatment recommendation domain \cite{wang2022hierarchical}. However, natural language as the primary communication media between people are rarely investigated in these methods. With the prevalence of transformer-based model and its capacity in addressing sequential data such as language, natural language is gradually included in training process which should become the bottleneck to enhance the explainability. 
In addition to intrinsically interpretable approaches, post-hoc explanation methodologies are prevalent in the research community \cite{xie2022towards,garg2022lisa,jiang2023interpretable}. Zhang et al. (2021) utilized a hierarchical framework to achieve post-hoc explanation, decomposing complex tasks and elucidating the model's decision-making process and causes of failure \cite{zhang2021explainable}. Wang et al. proposed subgoal-conditioned hierarchical IL to emulate doctors' behavior, providing more explainable post-hoc recommendations in the dynamic treatment recommendation domain \cite{wang2022hierarchical}. However, these methods seldom investigate natural language as the primary communication medium between individuals.
With the prevalence of transformer-based models and their capacity to address sequential data such as language, natural language is gradually being incorporated into the training process. This integration should serve as a bottleneck to enhance explainability, as it enables a more intuitive interaction between humans and machines.

%%%%%%%%%%%%%%%%%%%%%%%%%%%%%%%%%%%%%%%%%%%%%%%%%%%%%%%%%%%%%%%%%%%%%%%%

\section{Approach}
% summarize the subsection
% Ex-PERACT is a Explainable hierarchical agent for 6-DOF manipulation tasks, which trains in a behavior cloning fashion on a language-conditioned offline dataset. The key idea is extracting discrete skill codes to bridge the natural language instruction with the observation-action pairs in the higher level, and learn a policy network from the extracted skill code together with the observation-action pair in the lower level. The skill codes is learned unsupervisedly and enhance explainability with respect to the natural language and multitask learning. The 3D voxel reconstruction design is inherited from the PERACT and plays a salient role in learning efficiency compared with classical 2D image-action mapping.
Our approach, Ex-PERACT, introduces an explainable hierarchical agent for 6-DOF manipulation tasks. It adopts a behavior cloning approach, training on a language-conditioned offline dataset. The core concept involves extracting discrete skill codes to bridge natural language instructions with observation at a higher level, while simultaneously learning a policy network from the extracted skill code, language embeddings, and observation-action pair at a lower level. The skill codes is learned unsupervisedly and enhance explainability with respect to the natural language and multitask learning. Additionally, the design incorporates 3D voxel reconstruction inherited from PERACT, which significantly improves learning efficiency compared to conventional 2D image-action mapping.

% Section \ref{ps} introduces the preliminary knowledge and our problem formulation. Section \ref{method} provide more detail about our method Ex-PERACT. Further implementation is presented in Section \ref{imple}. 
We begin by presenting the preliminary knowledge and our problem formulation in Section \ref{ps}. Subsequently, Section \ref{method} delves into the framework of Ex-PERACT, providing a comprehensive introduction to the models. Finally, Section \ref{imple} elucidates further implementation details.

\subsection{Problem setting}\label{ps}
% task decomposing + MDP
% While we formulate our problem as multitask, each task $T_i$ could be viewed as individual Markov Decision Process, which satisfies the property that the next state $s_{t+1}$ is only determined by current state $s_t$ at any time $t$. 
% We assume access to an offline dataset $D$ generated by optimal policy, containing $m$ expert demonstration trajectories from a wide range of tasks $D = \{ \tau_1,\tau_2,...,\tau_m\}$. Each trajectory $\tau$ consists of a sequence of state-action pairs accompanied with the natural language instruction $\tau_i = \{l^i,\{ (s_1^i,a_1^i),(s_2^i,a_2^i),...,(s_n^i,a_n^i)\} \}$.
% For a given task $T_i$, it can be further decomposed into multiple snippets with various length, each snippet may represent a much simpler sub-goals. For example, task ``Take the steak off the grill" may decompose to sub-goals like ``Moving gripper to target the steak", ``picking up the steak", ``placing the steak to the destination". However, such kind of explicit instruction are uncommon in our daily life, people are used to ignore such kind of common sense. Our method Ex-PERACT tries to categorize these implicit steps at the top level model from the offline dataset.
% To simplify the problem, we assume the each snippet only encodes one sub-goal.
We conceptualize our problem as multitask, where each task $T_i$ can be regarded as an individual Markov Decision Process (MDP), characterized by the property that the next state $s_{t+1}$ is solely determined by the current state $s_t$ at any time $t$.
We assume access to an offline dataset $D$ generated by an optimal policy, comprising $m$ expert demonstration trajectories from a wide range of tasks, denoted as $D = {\tau_1,\tau_2,...,\tau_m}$. Each trajectory $\tau$ comprises a sequence of state-action pairs accompanied by a natural language instruction, structured as $\tau_i = {l^i,{(s_1^i,a_1^i),(s_2^i,a_2^i),...,(s_n^i,a_n^i)}}$.
Each task $T_i$ can be further decomposed into multiple snippets of various lengths, each representing simpler sub-goals. For instance, a task such as ``Take the steak off the grill" may decompose into sub-goals like ``Moving the gripper to target the steak", ``Picking up the steak", and ``Placing the steak at the destination". However, such explicit instructions are uncommon in daily life, often leading individuals to overlook common sense steps. Our method, Ex-PERACT, aims to categorize these implicit steps at the top-level model from the offline dataset, and to simplify the problem, we assume the each snippet only encodes one sub-goal.

% keyframes extraction + voxelization
% Like prior research, we follow the approach introduced by James et al. to decompose the trajectories \cite{james2022coarse}. A simple heuristic is designed to extract keyframes from demonstration: a timestep is chosen to be a keyframe if the joint velocity is close to zero or the gripper's status changed (from open to close, and vice versa). The actions of these extracted keyframes are also used as macro-action in training, which improves the robustness and overcome the randomness and noise caused by original continuous action.
In line with previous research, we adopt the approach introduced by James et al. for trajectory decomposition \cite{james2022coarse}. A simple heuristic is designed to extract keyframes from demonstration: a timestep is assigned as a keyframe if the joint velocity is close to zero or if the gripper's status changes (from open to close, and vice versa). The actions of these extracted keyframes are also utilized as macro-actions during training, enhancing robustness and mitigating the effects of randomness and noise inherent in original continuous actions.

\begin{figure}[ht]
\centering
\includegraphics[width=\linewidth]{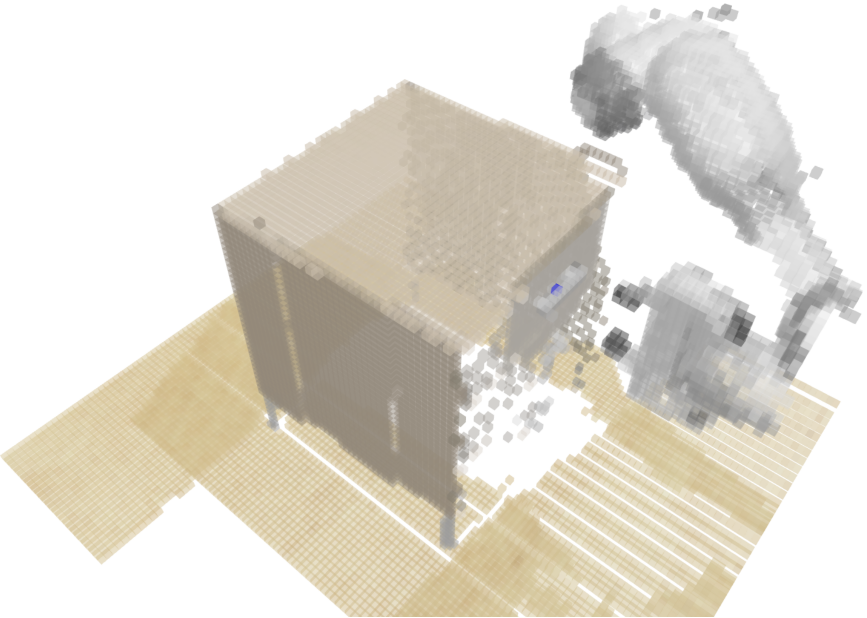}
\vspace{0.1cm}
\caption{Illustration of Voxelization for the ``Open Drawer" Task, adapted from \protect{\cite{shridhar2023perceiver}}. RGB-D image inputs undergo transformation into (100,100,100) voxel grids, representing a cubic meter space.}
\label{fig:voxel}
\vspace{0.6cm}
\end{figure}

% We follow PERACT's voxelization to reconstruct 3D representation from the RGB-D image observation \cite{shridhar2023perceiver}. This approach divides a $1.0m^3$ space into $100^3$ voxel grids (See Figure \ref{fig:voxel}). Such discrete representations of observation and action enable reformulating the problem as a "next best action" classification problem \cite{james2022coarse}, where the $(x,y,z)$ location of the voxel grid is the coordination of the center of gripper and the robotic arm's rotation is also discretised similarly over 3 axes with 5 degree resolution. 
We employ PERACT's voxelization method to reconstruct 3D representations from RGB-D image observations \cite{shridhar2023perceiver}. This approach partitions a $1.0m^3$ space into $100^3$ voxel grids (See Figure \ref{fig:voxel}). These discrete representations of observation and action facilitate reformulating the problem as a "next best action" classification problem \cite{james2022coarse}, wherein the $(x,y,z)$ location of the voxel grid corresponds to the coordination of the gripper's center, while the robotic arm's rotation is discretized similarly over three axes with a 5-degree resolution.

% no language-option label
% Note that although the keyframe extraction decomposes trajectories into snippets with various length, the sub-goals or skills represented by these snippets still left unknown. Due to the lack of this prior knowledge, we train top-level model unsupervisedly, using vector quantization to summarize skills into discrete skill code.
Despite decomposing trajectories into snippets of various lengths, the sub-goals or skills represented by these snippets remain unknown. Due to this lack of prior knowledge, we train the top-level model in an unsupervised manner, utilizing vector quantization to summarize skills into discrete skill codes.

\subsection{Ex-PERACT}\label{method}

\begin{figure*}[ht]
\centering
\includegraphics[width=\linewidth]{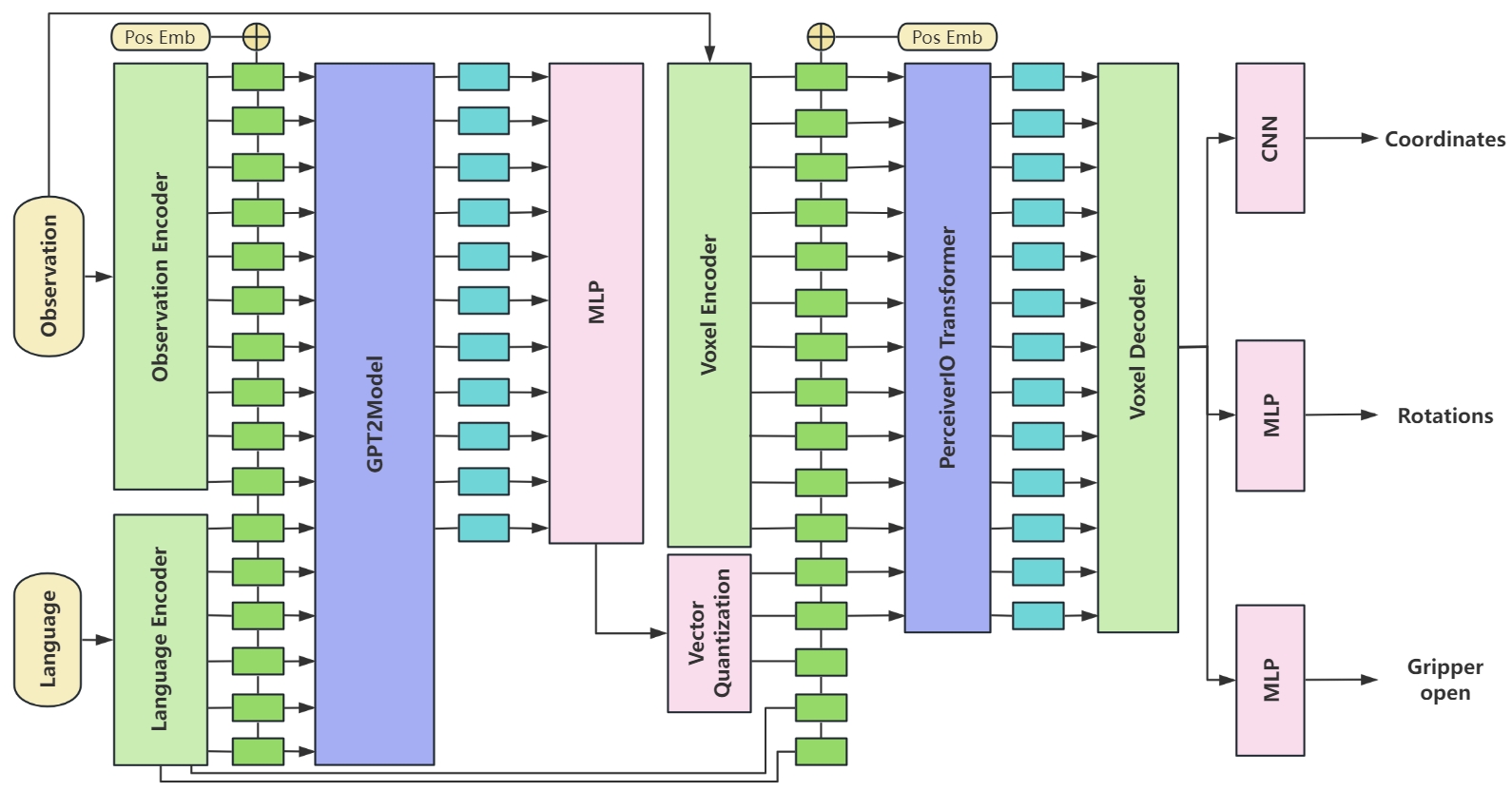}
\vspace{0.1cm}
\caption{A diagram illustrating the Ex-PERACT framework. Ex-PERACT, a hierarchical model, aims to learn reusable skills to improve explainability and predict actions based on given observations. Language instructions are encoded using a pre-trained CLIP model, while observations are preprocessed into embeddings or voxels. The top-level model takes language embeddings and observation embeddings as input, producing discretized skill code estimates. Meanwhile, the bottom-level model receives voxel inputs, language embeddings, and discretized skill code estimates, generating action predictions from three perspectives.}
\label{fig:framework}
\vspace{0.4cm}
\end{figure*}

%%%% Introduce the hierarchical models' I/O, encoding 
% As we mentioned before, Ex-PERACT is a two-level hierarchical framework which takes language instruction $\mathcal{L}$ and Observation $\mathcal{O}$ as input and outputs a discretised action including 3 aspects of behavior: coordination (also called translation), quaternion rotation, and gripper state $\mathcal{A} = \{\mathcal{A}_{coord},\mathcal{A}_{rotation},\mathcal{A}_{gripper}\}$ (see Figure \ref{fig:framework}). Both models all adopt the prevalent transformer-based approach to address the sequential input, the top level transformer learns discrete skill codes from a single language instruction and a sequence of observations $f(\mathcal{O},\mathcal{L}) \mapsto \mathcal{C}$; the bottom level transformer learns a policy in a command-conditional behavior cloning way $\pi(\mathcal{O},\mathcal{C}) \mapsto \mathcal{A}$, similar to \cite{codevilla2018end}. Motivated by promising result of CLIP for manipulation tasks in prior literature \cite{xiao2022robotic,shridhar2022cliport,guhur2023instruction}, Ex-PERACT also leverages CLIP \cite{radford2021learning}. CLIP is famous for its generalization and capacities in multimodal domain. 
% Tokenizer and pre-trained model of CLIP are loaded to tokenize and encode the human understandable language into language embedding with shape (77, 512), we further fine-tune the language embedding with a linear layer.

Ex-PERACT is a two-level hierarchical framework designed to process language instructions ($\mathcal{L}$) and observations ($\mathcal{O}$), yielding discretized actions comprising three aspects of behavior: coordination (also known as translation), quaternion rotation, and gripper state ($\mathcal{A} = {\mathcal{A}{\text{coord}},\mathcal{A}{\text{rotation}},\mathcal{A}_{\text{gripper}}}$) (see Figure \ref{fig:framework}). Both levels of the model employ transformer-based architectures to handle sequential input.
At the top level, a transformer model learns discrete skill codes from a single language instruction and a sequence of observations, represented as $f(\mathcal{O},\mathcal{L}) \mapsto \mathcal{C}$. Meanwhile, the bottom level transformer learns a policy in a command-conditional behavior cloning manner, denoted as $\pi(\mathcal{O},\mathcal{C},\mathcal{L}) \mapsto \mathcal{A}$, akin to previous work \cite{codevilla2018end}. 
Drawing inspiration from the promising results of CLIP in manipulation tasks reported in prior literature \cite{xiao2022robotic,shridhar2022cliport,guhur2023instruction}, Ex-PERACT also incorporates CLIP \cite{radford2021learning}. 
CLIP is renowned for its generalization capabilities and effectiveness in multimodal domains.
The tokenizer and pre-trained model of CLIP are employed to tokenize and encode human-understandable language into language embeddings with dimensions of (77, 512). Additionally, we fine-tune the language embeddings using a linear layer to further enhance their effectiveness.

The input observation consists of a set of observations, encompassing not only the observation at the current timestep but also those from both past and future keyframes. Each observation comprises RGB-D inputs from four cameras (front, left shoulder, right shoulder, and wrist).
As the heuristically identified keyframes divide each demonstration into multiple snippets, at timestep $t$, history observations are randomly sampled within the same snippet. The skill code at timestep $t$ is then obtained via majority vote from these history observations, ensuring consistency and robustness. Future keyframe observations are extracted from the remaining keyframes from timestep $t$.
All these observations are chronologically ordered and fed into the top-level model as a sequence. Initially, we fuse and encode the RGB-D input of each observation using a Convolutional Neural Network (CNN). The encoded observation embeddings are then combined with positional embeddings obtained from the timestep indexes of the observations and concatenated with the language embeddings from the pre-trained CLIP model. This language-observation embeddings are inputted into a simple causal transformer with a single self-attention layer followed by a linear layer, generating continuous skill code estimates for each observation in the input, i.e., $\hat{c}_t = f(o_t,l)$, where $o_t$ represents the individual observation at timestep $t$, and $l$ denotes the language instruction of the task. Due to the lack of prior knowledge regarding the skill codes, we utilize vector quantization $vq(\cdot)$ to cluster and discretize these learned continuous skill code estimates, yielding $\widetilde{c} = vq(\hat{c})$.

%%%% introduce the objective function, from the BC loss introduce the bottom model
% We train Ex-PERACT end-to-end under command-conditional behavior cloning paradigm, while standard command-conditional behavior cloning introduced in \cite{codevilla2018end} can be written as 
% \begin{equation}\label{eq:ccbc}
% minimize \sum_i \mathcal{L}(\pi(o_i,c_i),a_i).
% \end{equation}
% Since we have no prior knowledge to the skill codes and the skill codes are learned unsupervisedly in the top level model, we introduce vector quantization loss in equation \ref{eq:ccbc}. 
% % add how vector quantization loss is calculated
% Vector quantization loss, also called commitment loss, is calculated by the mean squared error between the discretised skill code estimate and the continuous skill code.
% At the mean time, remember the action is decomposed into 3 categories, we calculate the cross entropy loss of each type of action independently. In this case, the objective of Ex-PERACT could then be written as
% \begin{equation} \label{eq:experact}
% \begin{split}
% minimize \sum_i & \mathcal{L}_{CE}(\pi_{coord}(o_i,c_i),a_{i}^{\text{coord}}) + \\ & \mathcal{L}_{CE}(\pi_{rot}(o_i,c_i),a_{i}^{\text{rot}})+ \\ & \mathcal{L}_{CE}(\pi_{grip}(o_i,c_i),a_{i}^{\text{grip}})+ \\ & \mathcal{L}_{MSE}(f(o_i,l),\widetilde{c}_i),
% \end{split}
% \end{equation}
% where the last term could be directly obtain from the top level model, and the first 3 terms are calculated after each iteration.
We train Ex-PERACT end-to-end using the command-conditional behavior cloning paradigm. The standard command-conditional behavior cloning, as introduced in \cite{codevilla2018end}, can be represented by the following equation:
\begin{equation}\label{eq:ccbc}
\text{minimize} \sum_i \mathcal{L}(\pi(o_i,c_i),a_i).
\end{equation}
Since we lack prior knowledge of the skill codes, and they are learned unsupervisedly in the top-level model, we introduce vector quantization loss in Equation \ref{eq:ccbc}. Vector quantization loss is calculated as the mean squared error between the discretized skill code estimate and the continuous skill code. Additionally, considering the action decomposition into three categories, we independently calculate the cross-entropy loss for each type of action. Consequently, the objective of Ex-PERACT can be formulated as follows:
\begin{equation} \label{eq:experact}
\begin{split}
minimize \sum_i & \mathcal{L}_{CE}(\pi_{coord}(o_i,c_i,l),a_{i}^{\text{coord}}) + \\ & \mathcal{L}_{CE}(\pi_{rot}(o_i,c_i,l),a_{i}^{\text{rot}})+ \\ & \mathcal{L}_{CE}(\pi_{grip}(o_i,c_i,l),a_{i}^{\text{grip}})+ \\ & \mathcal{L}_{MSE}(f(o_i,l),\widetilde{c}_i),
\end{split}
\end{equation}
where the last term can be directly obtained from the top-level model, and the first three terms are calculated after each iteration.

To obtain various types of actions, we employ another transformer-based model to predict coordinates, rotation, and gripper state separately. At timestep $t$, the top-level model takes the observation set and language instruction as input, while at the bottom level, the same observation set and language embeddings together with obtained skill code estimates are used as input. The observations are reconstructed into voxels $v$ with a shape of $(100,100,100)$ using a similar approach as described in both \cite{james2022coarse} and \cite{shridhar2023perceiver}. These $100^3$ voxel grids are then split and flattened into sequences with a length of $(100/5)^3=8000$ using a 3D convolutional layer with kernel and stride size set to five. The flattened voxel sequence is subsequently concatenated with the extracted skill code estimate and language embeddings, and then fed into a six-layer self-attention model.
Considering the potential memory limitations when processing long sequences on commodity GPUs, we adopt the Perceiver Transformer $pt(\cdot)$ with 512 latents of dimension 512 \cite{jaegle2021perceiver}. The Perceiver Transformer allows for projection from a long sequence to a much shorter latent space and then projects the final output latent back to the original input size. The features of each voxel grid are obtained from the output of the Perceiver Transformer after decoding, which are then utilized to predict coordinates using a 3D CNN or rotation and gripper status using a Multilayer Perceptron (MLP). Specifically:
\begin{equation} \label{eq:bottom}
\begin{split}
\hat{a}_{t}^{\text{coord}}  = & \pi_{coord}(pt(o_t,c_t,l))\\
\hat{a}_{t}^{\text{rot}}  = & \pi_{rot}(pt(o_t,c_t,l))\\
\hat{a}_{t}^{\text{grip}}  = & \pi_{grip}(pt(o_t,c_t,l)).
\end{split}
\end{equation}

%describe the table
\begin{table*}[ht]
\caption{Multitask performance comparison across state-of-the-art methods. Values represent mean success rate (\%) obtained from 25 evaluation episodes per task with the best ones bolded.}
\vspace{0.3cm}
\centering
\label{t:multitask_p}
\begin{tabular}{@{}ccccccccc@{}}
\toprule
method & open drawer & meat off grill & slide block & turn tap & close jar & stack blocks & screw bulb & push buttons \\ \midrule
BC (CNN) & 4 & 0 & 4 & 20 & 0 & 0 & 0 & 4 \\
BC (ViT) & 16 & 0 & 8 & 24 & 0 & 0 & 0 & 16 \\
LISA & 8 & 0 & 4 & 28 & 0 & 0 & 0 & 20 \\
C2FARM-BC & 28 & 40 & 12 & 60 & 28 & 4 & 12 & 88 \\
Ex-PERACT (Obs only) & 20 & 40 & 8 & 36 & 16 & 0 & 0 & 60 \\
Ex-PERACT (Obs + skill code) & 28 & 56 & 16 & 48 & 16 & 4 & 20 & 72 \\
\begin{tabular}[c]{@{}c@{}}Ex-PERACT \\ (Obs + Lang, which is PERACT)\end{tabular} & 64 & 68 & 32 & 60 & 28 & \textbf{8} & 28 & 56 \\
\rowcolor[HTML]{CFFECE} 
Ex-PERACT (single Lang) &  \textbf{76}& \textbf{80} & \textbf{44} & 60 & 32 & 0 & \textbf{36} & \textbf{92} \\
\rowcolor[HTML]{CFFECE} 
Ex-PERACT (multi Lang) &  72& \textbf{80} & \textbf{44} & \textbf{88} & \textbf{40} & \textbf{8} & 24 & 88 \\
\bottomrule
\end{tabular}
\vspace{0.4cm}
\end{table*}

\subsection{Implementation}\label{imple}
The Ex-PERACT framework is hierarchical in nature, and we employ the single LAMB optimizer \cite{you2019large} to update parameters from both the top and bottom-level models. To facilitate end-to-end training within a framework containing non-differentiable components like VQ, we utilize a technique known as the straight-through estimator. This method directly propagates gradients from the decoder to the encoder.
Given that actions are discretized, cross-entropy losses for three types of actions are computed by comparing the action estimates encoded with ground truth one-hot encoding. Specifically, for rotation, the loss is calculated under the 3D Euler angle coordinate setting, with the final rotation action reconstructed from the Euler angles into a quaternion.
Demonstration inputs are obtained with the assistance of a motion planner in RLBench, with each demonstration sourced from a different variance number. Since observation inputs for specific timesteps consist of observation sets, we set the sample frequency and batch size to 10 and 1, respectively, to ensure efficient disk and memory usage on commodity devices.
Observations reveal significant variations in demonstration length across tasks, ranging from approximately 100 to over 400 timesteps. Additionally, the number of identified keyframes exhibits considerable deviation, ranging from 2 to around 20. To maintain a similar sample frequency for each sample, we set the sampling weights as the sum of the total number of observations for each task and the size of the observation set.
In the bottom-level model, we implement data augmentation prior to voxelization.

\section{Result}
% introduce the summary of this section, 
% In this section, we introduce our experiment settings and results. Ex-PERACT is examined under RLBench manipulation tasks. We compare its multitasks policy performance with other state-of-the-art methods, and analyse the explainability brings from the hierarchical framework.
In this section, we present our experimental setup and findings. We assess Ex-PERACT's performance using RLBench manipulation tasks, comparing its multitask capabilities against those of other state-of-the-art methods. Additionally, we analyze the explainability provided by the hierarchical framework.

\subsection{Experiment setup}
Ex-PERACT is evaluated within the RLBench simulation platform, utilizing a GPU NVIDIA Quadro RTX 5000 for computations. RLBench operates on CoppeliaSim/V-REP \cite{rohmer2013v} and PyRep \cite{james2019pyrep} interfaces, featuring a Franka Emika Panda robot arm with six degrees of freedom (DOF). This platform provides crucial data such as joint velocities, angles, and forces, alongside four $128 \times 128$ RGB-D images captured from different camera perspectives. Diverse manipulation tasks are designed, ensuring variability in task execution and accompanying language summaries. To manage memory usage effectively, all samples are stored on disk.
We select eight tasks spanning from opening a drawer to screwing in a light bulb, with ten demonstrations collected for each task. For comparison, we evaluate several state-of-the-art imitation learning (IL) methods as baselines. Baseline agents include BC with image encoder CNN and Vision Transformer (ViT) \cite{dosovitskiy2020image}, LISA \cite{garg2022lisa}, C2FARM-BC \cite{james2022coarse}, and PERACT \cite{shridhar2023perceiver}. BC and LISA are image-to-action agents mapping observed RGB-D images to eight-dimensional actions. In contrast, C2FARM-BC, PERACT, and Ex-PERACT are voxel-to-action agents, reconstructing the 3D space and discretizing the problem as a classification task. While C2FARM-BC adopts a two-level voxelization allowing zoom-in operations for higher resolution ($0.47cm^3$), PERACT and Ex-PERACT employ a voxel resolution of $1cm^3$.
Each agent interacts with the test environment via a motion planner, and the mean success rate is computed from 25 episodes per task ($8\times25=200$ total episodes evaluated). Success is defined as reaching the goal state within 25 steps, while episodes encountering issues such as exceeding the maximum steps, encountering an invalid path, or having the target outside the workspace are deemed failures.

\subsection{Policy performance}

Table \ref{t:multitask_p} presents the average success rates of state-of-the-art methods and our method Ex-PERACT. The first three methods are image-based BC agents, while the remaining methods employ a 3D representation. The ``single lang" Ex-PERACT denotes that the language instructions are in a consistent format for a specific task, whereas the ``multi-lang" Ex-PERACT indicates variable instruction formats across batches. For example, in the task ``open drawer", a consistent instruction would be ``open the top/middle/bottom drawer"; under the ``multi-lang" condition, the instructions could vary randomly, including ``open the top/middle/bottom drawer", ``grip the top/middle/bottom handle and pull the top/middle/bottom drawer open", or ``slide the top/middle/bottom drawer open". All methods receive the same set of 80 demonstrations ($8\times10=80$) as input. Analysis of Table \ref{t:multitask_p} reveals that image-based methods demonstrate poor performance across all tasks, whereas those utilizing 3D representation consistently outperform them. With only 10 demonstrations available per task, image-based agents face significant challenges across all eight tasks. This finding underscores the formidable task of learning hand-eye coordination from scratch, given its inherent demands for time and data. Furthermore, this comparison supports our contention that 3D voxelization offers enhanced data efficiency. Among methods employing 3D representation, our approach, Ex-PERACT, which integrates various language inputs, consistently demonstrates superior policy performance. It surpasses PERACT by an average improvement of 29.07\% (the third row from the bottom) and achieves a 63.23\% better average performance compared to C2FARM. 

We observe that C2FARM, PERACT, and Ex-PERACT perform better in tasks like ``open drawer", ``meat off grill", and ``turn tap", which have relatively fewer variations, namely 3 (top/middle/bottom), 2 (steak/drumstick), and 2 (left/right), respectively. This pattern indicates that the performance of BC agents is highly dependent on the number of variations in a task. Moreover, all evaluated methods show limited success in the task ``stack blocks" (with a maximum of 60 variations). In addition to the influence of highest variation number, we think the principal challenge lies in the degree of freedom of the manipulable objects. The more states these objects can reach, the more challenging it is for BC agents to learn, as minor errors can be magnified during the manipulation of multiple objects. For example, although Ex-PERACT can successfully stack the first block, any minor misalignment in the location or rotation of the first blocks degrades the performance of subsequent actions. The impact of such errors becomes even amplified as more blocks are stacked. This is likely a major contributing factor to the difficulties encountered in the ``stack blocks" task, considering the promising performance in task ``push buttons" while ``push buttons also possesses a maximum 50 variation number.

We conducted an ablation study to investigate the impact of skill code availability, language, and their combination, as well as language diversity, on performance. The results of this analysis are presented in the fifth-to-last row of Table \ref{t:multitask_p}. Comparing Ex-PERACT with observation only to Ex-PERACT with both skill code and observation, we conclude that the inclusion of skill code provides useful information to enhance agent policy performance. In comparing Ex-PERACT with skill code to Ex-PERACT with language (PERACT), we observed that Ex-PERACT with language significantly outperforms Ex-PERACT with skill code. This finding contradicts the result in \cite{garg2022lisa}, where non-hierarchical Ex-PERACT with language performed worse than hierarchical Ex-PERACT with skill code. We believe the reason for this discrepancy is that the limited number of skills is inadequate to represent tasks with large varieties. This assertion is supported by the performance of Ex-PERACT with skill code, particularly in low variety tasks such as ``meat off grill" and ``turn tap", where the agent performs relatively better than in tasks with greater variety. Additionally, comparing the performance between single-language Ex-PERACT and multi-language Ex-PERACT, the results indicate that the inclusion of diverse language instructions does not detract from the performance of Ex-PERACT.

\subsection{Explainable skill representation}
%Explain the image
% To examine the explainability of Ex-PERACT, we record the input language instructions together with the output skill codes in a dictionary. The intuition is using the words with high frequency to describe the learned skill code. After trying different visualization methods, we found that it is less intuitive to use the traditional visualization approach such as heat map to present the skill codes' verbal meaning, since various format of language instruction is included and the large number of available words will make the heat map overwhelming. In this case, we decide to use word cloud to enhance the accessibility for less technical audience. 
To evaluate the explainability of Ex-PERACT, we aggregated the input language instructions alongside the corresponding output skill codes into a dictionary. In this dictionary, each key represents a skill code generated by the Ex-PERACT system, while the corresponding value comprises a list of words extracted from the language instructions. Notably, common stop words such as ``from", ``it", and ``to" were excluded from the list. The rationale behind this methodology is to leverage frequently occurring words to elucidate the learned skill codes. Following this compilation, we explored various visualization techniques to represent the human-understandable verbal meanings of the skill codes. Traditional methods, such as heat maps, are less intuitive due to the heterogeneous formats of language instructions and the potential inundation of words, resulting in a dense and overwhelming visualization. Consequently, we opted for a word cloud to enhance accessibility for less technically inclined audiences.

\begin{figure}[t]
     \begin{subfigure}{0.24\textwidth}
         \includegraphics[width=\textwidth]{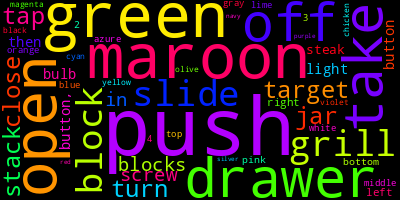}
         \caption{Word cloud for skill code 04.}
         \label{wordcloud_04}
         \vspace{0.3cm}
     \end{subfigure}
     \begin{subfigure}{0.24\textwidth}
         \includegraphics[width=\textwidth]{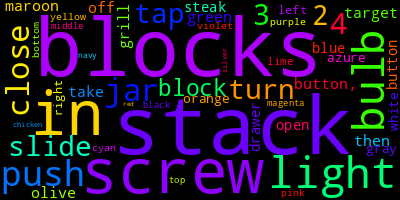}
         \caption{Word cloud for skill code 10.}
         \label{wordcloud_10}
         \vspace{0.3cm}
     \end{subfigure}
     \begin{subfigure}{0.24\textwidth}
         \includegraphics[width=\textwidth]{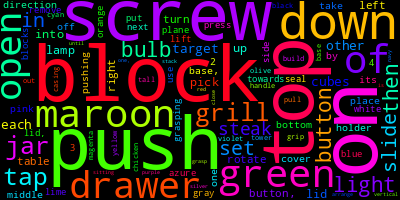}
         \caption{Word cloud for skill code 18.}
         \label{wordcloud_18}
         \vspace{0.5cm}
     \end{subfigure}
     \begin{subfigure}{0.24\textwidth}
         \includegraphics[width=\textwidth]{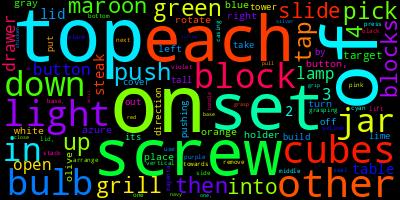}
         \caption{Word cloud for skill code 19.}
         \label{wordcloud_19}
         \vspace{0.5cm}
     \end{subfigure}
     \caption{Word Clouds for different skill codes, with the size of each word indicating its frequency within the corresponding skill code.}
     \label{exp}
     \vspace{0.5cm}
\end{figure}
Figure \ref{exp} showcases several word cloud examples corresponding to different skill codes. Upon inspection, it becomes evident that each skill code emphasizes distinct aspects of the evaluated tasks. For instance, Skill Code 04 (refer to Figure \ref{wordcloud_04}) appears to hold greater significance in tasks such as ``open drawer", ``meat off grill", and "slide block", with a notable focus on colors. Conversely, Skill Code 10 (refer to Figure \ref{wordcloud_10}) is centered around tasks like "stack blocks" and ``screw bulb", placing emphasis on numerical values to indicate the quantity of objects being manipulated.
Furthermore, Skill Codes 18 and 19, both associated with the action of ``screw", but exhibit unique characteristics (refer to Figures \ref{wordcloud_18} and \ref{wordcloud_19}). Skill Code 18 demonstrates a closer relationship to the task ``slide block", featuring words like ``tap" and ``jar" that denote screw-like actions, albeit without explicit mentions of ``screwing". In contrast, Skill Code 19 appears to highlight the relative positions of manipulable objects.

By examining the common tasks associated with each skill code, we can derive abstract insights into the underlying sub-actions represented by the skills. For example, Skill Code 04 could suggest a common sub-action of ``grabbing an object" across tasks like ``open drawer", ``meat off grill", and ``slide block". Similarly, Skill Code 10 implies a common sub-action of ``placing an object on top of another" in tasks such as ``stack blocks" and ``screw bulb". However, it's important to note that these interpretations are based on human induction, as there may be a lack of alignment between the language instructions and the delineation of key points.Future endeavors could focus on refining this process by decomposing the language instructions and correlating language fragments with demonstration snippets to generate more convincing explanations.

\section{Conclusion and Future Work}
% In this paper, we proposed Ex-PERACT, an explainable multi-task behavior cloning agent with hierarchical structure, the top-level is used to learn discrete skills, and the bottom-level leverages skills, language, and observation to learn proper behaving policy. Our experiments on 6-DOF manipulation tasks shows that Ex-PERACT demonstrates its promising performance compared to other baselines while providing an interpretable visualization on language instruction and skills learned by agent. 

This paper introduces Ex-PERACT, an explainable hierarchical multi-task behavior cloning agent. At the top level, Ex-PERACT learns discrete skill codes via clustering, while the bottom level utilizes skill codes, language, and observations to acquire a proficient behavioral policy by representing the problem as discrete 3D voxel grids. Our experiments on 6-DOF manipulation tasks illustrate that Ex-PERACT demonstrates promising performance compared to other baseline methods, concurrently providing an human-understandable visualization of skills learned by the agent.

However, certain limitations of Ex-PERACT persist and warrant exploration as potential future research directions. Given Ex-PERACT's ability to accommodate various forms of language instructions, developing a model to decompose various language instructions to cooperate with trajectory key points could be one interesting direction to further enhance explainability. Additionally, Ex-PERACT faces challenges in long-term tasks and tasks characterized by high variability. Subsequent research endeavors may prioritize enhancing its performance, addressing not only the breadth of tasks but also considerations of task length and variance.

%%%%%%%%%%%%%%%%%%%%%%%%%%%%%%%%%%%%%%%%%%%%%%%%%%%%%%%%%%%%%%%%%%%%%%%%

%%%%%%%%%%%%%%%%%%%%%%%%%%%%%%%%%%%%%%%%%%%%%%%%%%%%%%%%%%%%%%%%%%%%%%%%

%%% Use this environment to include acknowledgements (optional).
%%% This will be omitted in doubleblind mode.

%%%%%%%%%%%%%%%%%%%%%%%%%%%%%%%%%%%%%%%%%%%%%%%%%%%%%%%%%%%%%%%%%%%%%%%%

%%% Use this command to include your bibliography file.

\bibliography{main}

\end{document}